\documentclass[letterpaper]{article} % DO NOT CHANGE THIS
\usepackage{aaai2026}  % DO NOT CHANGE THIS
\usepackage{times}  % DO NOT CHANGE THIS
\usepackage{helvet}  % DO NOT CHANGE THIS
\usepackage{courier}  % DO NOT CHANGE THIS
\usepackage[hyphens]{url}  % DO NOT CHANGE THIS
\usepackage{graphicx} % DO NOT CHANGE THIS
\usepackage{natbib}  % DO NOT CHANGE THIS AND DO NOT ADD ANY OPTIONS TO IT
\usepackage{caption} % DO NOT CHANGE THIS AND DO NOT ADD ANY OPTIONS TO IT
\usepackage{algorithm}
\usepackage{algorithmic}

\usepackage{multirow}
\usepackage{amsfonts}
\usepackage{mathtools}
\newtheorem{theorem}{Statement}[section]
\DeclarePairedDelimiterX{\inp}[2]{\langle}{\rangle}{#1, #2}

\title{FunKAN: Functional Kolmogorov-Arnold Network for Medical Image Enhancement and Segmentation}
\author{
    Maksim Penkin\textsuperscript{\rm 1},
    Andrey Krylov\textsuperscript{\rm 1}
}
\affiliations{
    \textsuperscript{\rm 1}Laboratory of Mathematical Methods of Image Processing, Faculty of Computational Mathematics and Cybernetics, Lomonosov Moscow State University\\
    1-52, Leninskiye Gory\\
    Moscow, 119991 Russia\\
    penkin97@gmail.com, kryl@cs.msu.ru
}

\begin{document}
\maketitle

\begin{abstract}
Medical image enhancement and segmentation are critical yet challenging tasks in modern clinical practice,
constrained by artifacts and complex anatomical variations.
Traditional deep learning approaches often rely on complex architectures with limited interpretability.
While Kolmogorov-Arnold networks offer interpretable solutions,
their reliance on flattened feature representations fundamentally disrupts the intrinsic spatial structure of imaging data.
To address this issue we propose a Functional Kolmogorov-Arnold Network (FunKAN)
-- a novel interpretable neural framework, designed specifically for image processing, that
formally generalizes the Kolmogorov-Arnold representation theorem onto functional spaces
and learns inner functions using Fourier decomposition over the basis Hermite functions.
We explore FunKAN on several medical image processing tasks, including Gibbs ringing suppression in magnetic resonance images,
benchmarking on IXI dataset.
We also propose U-FunKAN as state-of-the-art binary medical segmentation model with benchmarks on three medical datasets:
BUSI (ultrasound images),
GlaS (histological structures)
and CVC-ClinicDB (colonoscopy videos), detecting breast cancer, glands and polyps, respectively.
Experiments on those diverse datasets demonstrate that our approach outperforms other KAN-based backbones
in both medical image enhancement (PSNR, TV) and segmentation (IoU, F1).
Our work bridges the gap between theoretical function approximation and medical image analysis,
offering a robust, interpretable solution for clinical applications.
\end{abstract}

\begin{links}
    \link{Code}{https://github.com/MaksimPenkin/MedicalKAN}
\end{links}

\section{Introduction}
Computer-aided diagnosis systems~\cite{kadhim2022deep}
now constitute essential components of the contemporary medical imaging infrastructure,
effectively mitigating key challenges, including escalating diagnostic volumes and minimizing interpretation errors.
Still, persistent issues in image quality and segmentation accuracy hinder optimal clinical applications.

Deep learning has transformed the field of medical image analysis, however,
neural architectures often employ empirically designed components  
that lack theoretical foundations and struggle with modality-specific variations~\cite{borys2023explainable}.
Thus, this research specifically targets two crucial medical imaging needs:
first, developing theoretically grounded architectures
for magnetic resonance imaging (MRI) enhancement,
addressing Gibbs artifacts, often coupled with blur and noise;
second, creating new medical segmentation AI-tools for cancer detection in diverse imaging contexts.

MRI utilizes magnetic field gradients~\cite{epstein2007introduction} to encode spatial data in the frequency domain,
creating fine-resolved images of anatomical details that would normally be obscured by wavelength restrictions.
However, constraints of scan time and signal-to-noise level limit the amount of the k-space data that can be sampled. 
Once the k-space data have been acquired,
spatial decoding is performed
typically by inverse Fourier transform,
which leads to Gibbs ringing, also known as truncation artifact~\cite{wilton1928gibbs}.
It occurs, as the Fourier series cannot represent a discontinuity
with a finite number of harmonics~\cite{mallat1999wavelet}.
So, the oscillations may be observed around the edges of axial brain images
due to the signal intensity difference at locations,
such as the CSF-spinal cord or the skull-brain interface.

Extensive clinical evidence confirms that early-stage disease detection and subsequent diagnostic confirmation,
whether through ultrasound imaging, histopathological analysis or colonoscopy,
correlate strongly with enhanced long-term survival probabilities~\cite{abhisheka2023comprehensive}.
Diagnostic accuracy, however, is compromised by an exponential increase in imaging examinations
coupled with a critical shortage of trained specialists, including radiologists and pathologists.
Indeed, the World Health Organization (WHO) reported that in 2024
breast cancer affected 2.3 million women worldwide annually, resulting in 670000 deaths.
This disease can develop at any age after puberty.
Breast cancer outcomes exhibit a significant association with socioeconomic development levels:
in very high-HDI nations women face a lifetime breast cancer incidence of 1 in 12 and a mortality rate of 1 in 71.
In stark contrast, low-HDI countries demonstrate both a lower incidence rate of 1 in 27
and disproportionately higher mortality 1 in 48,
highlighting the substantial deficiencies in early detection capabilities and therapeutic accessibility.

In light of the advancements of Kolmogorov-Arnold networks (KANs)~\cite{liu2024kan}
for both medical image enhancement~\cite{penkin2025adaptive} and segmentation~\cite{li2025u},
we propose a Functional Kolmogorov-Arnold Network (FunKAN) --
a novel extension of the original KANs to better address fundamental image processing requirements.
While the original theorem~\cite{kolmogorov1957representations} applies to continuous functionals
$f(x_{1}, ..., x_{n})$ on $\mathbb{R}^{n}$, 
we hypothesize its generalization to continuous functionals
$f(\chi_{1}, ..., \chi_{n})$ on $H^{n}$,
where each $\chi_{i}$ states for an element from a Hilbert space $H$.
This extension remains formally unproven but, if valid, would significantly expand the theorem's applicability,
providing the mathematical basis for seamless incorporation of Kolmogorov-Arnold networks as modular components
within convolutional neural networks (CNNs) -- prevailing architectures for image processing tasks.
The proposed functional extension enables KAN-based feature extraction in a natural way for high-dimensional feature spaces,
considering each 2D feature map $\chi_{i}$ as an element of an underlying Hilbert space $H$,
viewed on a spatial grid $h \times w$.
Thus, the proposed approach preserves an intrinsic structure of imaging data by obviating feature flattening
and establishes a principled connection between the classical approximation theory and the contemporary deep learning approaches for image analysis.

Our contributions can be summarized as follows:
\begin{enumerate}
\item \textbf{Theoretical contribution:} We propose an extension of the Kolmogorov-Arnold theorem onto functional spaces.

\item \textbf{Empirical validation:} We introduce Functional Kolmogorov-Arnold Network (FunKAN)
and empirically validate it through comprehensive experiments on medical image enhancement and segmentation tasks.

\item \textbf{MRI enhancement superiority:} FunKAN outperforms existing KAN-based backbones, including
spline-based KAN~\cite{liu2024kan},
ChebyKAN~\cite{ss2024chebyshev}
and HermiteKAN~\cite{seydi2024exploring},
demonstrating superior MRI enhancement capabilities on IXI dataset~\cite{zhao2020gibbs}.

\item \textbf{Medical segmentation benchmarking:} U-FunKAN achieves state-of-the-art segmentation accuracy across three distinct medical imaging modalities:
\begin{itemize}
    \item breast ultrasound, BUSI dataset~\cite{al2020dataset},
    \item histological gland structures, GlaS dataset~\cite{valanarasu2021medical},
    \item colonoscopy polyp detection, CVC-ClinicDB dataset~\cite{bernal2015wm}.
\end{itemize}

\item \textbf{Reproducible research:} We release entire codebase on GitHub featuring:
\begin{itemize}
    \item PyTorch Lightning for modularity,
    \item Ruff for code quality enforcement,
    \item YAML-based configuration system for experiments management.
\end{itemize}
\end{enumerate}

\section{Related Work}
\subsection{Magnetic Resonance Image Enhancement}
Numerous computational approaches have been proposed for mitigating Gibbs artifacts in magnetic resonance imaging,
falling into three principal categories:
window-based filtering techniques,
total variation minimization methods 
and deep learning approaches, with some algorithms combining these multiple paradigms into hybrid frameworks.

Early approaches~\cite{veraart2016gibbs} include the Lanczos local averaging method
and the use of the Hann window, also known as the raised cosine window,
in fast Fourier transform (FFT) to smooth discontinuities.
Hanning window and other popular windowing functions mitigate the effects, caused by discontinuities,
that occur when applying FFT to finite-length signals. However, in practice, such methods often lead to blurry images.

Later on, another way~\cite{kellner2016gibbs} to mitigate Gibbs ringing was introduced 
through optimal subvoxel shifts search that minimize total variation:
\begin{equation} \label{eq:total_var}
TV(I,D) = \int_{D} \lvert \nabla{I(x)} \rvert dx,
\end{equation}
where $I$ is an image, defined on a domain $D$ (e.g. unit square $[0, 1]^{2}$).
The key idea is to acquire multiple versions of the image with slight shifts (fractions of a voxel)
and combine them in a way that cancels out the oscillatory ringing while preserving the edge sharpness.
In particular, the Kellner algorithm expresses an image $I$ with the Fourier series 
and derives subvoxel shifts from the FFT shift theorem.

Recent developments in deep learning have yielded sophisticated neural networks for Gibbs artifact suppression in MRI. 
The contemporary approaches predominantly employ convolutional neural networks to learn nonlinear mappings
between artifact-corrupted and artifact-free MR images~\cite{zhao2020gibbs}.
This paradigm also demonstrates particular effectiveness
through the hybrid architectures~\cite{penkin2021hybrid, penkin2023medical},
providing efficient post-processing solutions that combine computational speed with artifact reduction performance.

\subsection{Medical Image Segmentation}
Deep learning has also driven a substantial progress in medical image segmentation,
enabling automated and precise delineation of anatomical structures.

U-Net~\cite{ronneberger2015u} established a foundational encoder-decoder framework with skip connections,
facilitating an accurate localization through the integration of high-level semantic information with low-level spatial details.
Its widespread adoption in medical image analysis is largely attributable to its robust performance with limited training data.
However, the original U-Net architecture exhibits limitations in modeling long-range spatial dependencies and preserving fine structural details.

Early architectural innovations addressed these limitations through several key developments.
Attention U-Net~\cite{oktay2018attention} enhances feature selectivity through attention gates in skip connections,
dynamically emphasizing salient features, useful for a specific task, while suppressing irrelevant ones.
U-Net++~\cite{zhou2018unet++} improves feature fusion through nested, dense skip pathways,
reducing the semantic gap between encoder and decoder features.
By aggregating features across multiple scales U-Net++
enhances segmentation quality for anatomically irregular targets
(e.g., infiltrating tumor margins),
while incurring a greater computational overhead.

Emerging hybrid architectures have introduced novel computational paradigms,
like U-Mamba~\cite{ma2024u} and U-KAN~\cite{li2025u}.
U-Mamba integrates Mamba into the U-Net architecture to capture long-range dependencies
with linear computational complexity, making it particularly suitable for high-resolution medical imaging.
U-KAN substitutes convolutional layers in the backbone with Kolmogorov-Arnold-motivated adaptive activation functions.
The spline-based parameterization of these activations enables more accurate modeling of the complex biological morphologies,
particularly irregular tumor margins and vascular networks,
while simultaneously addressing the spectral bias~\cite{rahaman2019spectral} inherent in ReLU-based neural networks.
However, the proposed KAN-based backbone processes spatial feature maps as unstructured coordinate collections,
thereby ignoring the locality priors essential for image representation.
MedKAN~\cite{yang2025medkan} and UKAGNet~\cite{drokin2024kolmogorov} partially mitigate this limitation 
through a hybrid convolutional approach, combining adaptive spline-based nonlinearities with spatial inductive biases.
However, MedKAN remains constrained by its classification design,
lacking any proven generalizations onto medical image enhancement and segmentation pipelines.
Whereas UKAGNet does not go beyond the original Kolmogorov-Arnold theorem to adapt the concept
further for image-to-image processing pipelines.

\subsection{Kolmogorov-Arnold Networks}
Cutting-edge deep learning research is increasingly grounded in rigorous mathematical foundations,
enabling advanced modeling of complex data relationships~\cite{li2023fourier}.
A prime example is Kolmogorov-Arnold network~\cite{liu2024kan},
which implements the theoretical framework of the Kolmogorov-Arnold theorem through adaptive B-spline embeddings.

The Kolmogorov-Arnold theorem~\cite{kolmogorov1957representations} states that if
$f$: $[0, 1]^{n} \rightarrow \mathbb{R}$ is a multivariate continuous function, then
it can be written as a finite composition of continuous functions of a single variable and the binary operation of addition:
\begin{equation} \label{eq:kan_theorem}
  f(x_{1}, ..., x_{n}) = \sum_{j=1}^{2n+1} \zeta_{j} (\sum_{i=1}^{n} \phi_{ji}(x_{i})),
\end{equation}
where
$\phi_{ji}$: $[0, 1] \rightarrow \mathbb{R}$ and
$\zeta_{j}$: $\mathbb{R} \rightarrow \mathbb{R}$ -- continuous inner functions of a single variable.
The theorem provides another justification that neural networks of sufficient depth and width
are capable of forming dense subsets
in the space of continuous functions defined over compact domains~\cite{cybenko1989approximation}.

While the Kolmogorov-Arnold representation theorem offered a theoretically appealing reduction
of high-dimensional function approximation to learning univariate functions,
the pathological non-smoothness or even fractal character of its inner functions had severely constrained
its applications.
In 2024 the authors~\cite{liu2024kan} presented Kolmogorov-Arnold network,
relaxing the original theorem's constraints while preserving its fundamental principles.
Unlike the classical representation limited to two nonlinear layers with $(2n + 1)$ hidden terms,
their architecture permits arbitrary width and depth, leading to the modern differentiable KAN definition:
\begin{equation} \label{eq:kan}
  KAN(\mathbf{x}) = (\Phi_{L} \circ \Phi_{L-1} \circ ... \circ \Phi_{1})(\mathbf{x}),
\end{equation}
where $\{ \Phi_{l} \}_{l=1}^{L}$ -- the Kolmogorov-Arnold layers, defined as:
\begin{equation} \label{eq:kan_layer_matrix}
\begin{bmatrix}
    x_{l+1,1} \\
    x_{l+1,2} \\
    \vdots  \\
    x_{l+1,m}
\end{bmatrix}
=
\begin{bmatrix}
    \phi_{l,11} & \phi_{l,12} & ...    & \phi_{l,1n} \\
    \phi_{l,21} & \phi_{l,22} & ...    & \phi_{l,2n} \\
    \vdots      & \vdots      & \vdots & \vdots      \\
    \phi_{l,m1} & \phi_{l,m2} & ...    & \phi_{l,mn} 
\end{bmatrix}
\begin{bmatrix}
    x_{l,1} \\
    x_{l,2} \\
    \vdots  \\
    x_{l,n}
\end{bmatrix},
\end{equation}
\begin{equation} \label{eq:kan_layer}
x_{l+1,j} = \sum_{i=1}^{n} \phi_{l,ji}(x_{l,i}),
\end{equation}
where $\phi_{l,ji}$: $\mathbb{R} \rightarrow \mathbb{R}$ -- continuous inner functions, parameterized in a smooth differentiable manner by B-splines.

Recent KAN architectures have improved efficiency
by replacing B-splines with Gaussian radial basis functions (RBFs), resulting in FastKAN~\cite{li2024kolmogorov}.
In ChebyKAN~\cite{ss2024chebyshev} the authors employ Chebyshev polynomials as a complete orthogonal system to substitute B-splines,
achieving enhanced training stability. 

Although Kolmogorov-Arnold networks demonstrate powerful capabilities in multivariate function approximation,
their naive application to image processing is fundamentally limited.
The limitation stems from KANs' treatment of inputs as permutation-invariant scalars, 
thereby ignoring the essential two-dimensional geometric structure inherent in visual data.
We resolve this limitation by formulating a functional-space generalization of the Kolmogorov-Arnold theorem,
considering each 2D feature map as an element of an underlying Hilbert space $H$.
This theoretical advancement motivates our functional Kolmogorov-Arnold network,
seamlessly suitable for image processing pipelines by incorporating spatial awareness.

\section{Method}
\begin{figure*}
\centering
\includegraphics[width=\textwidth]{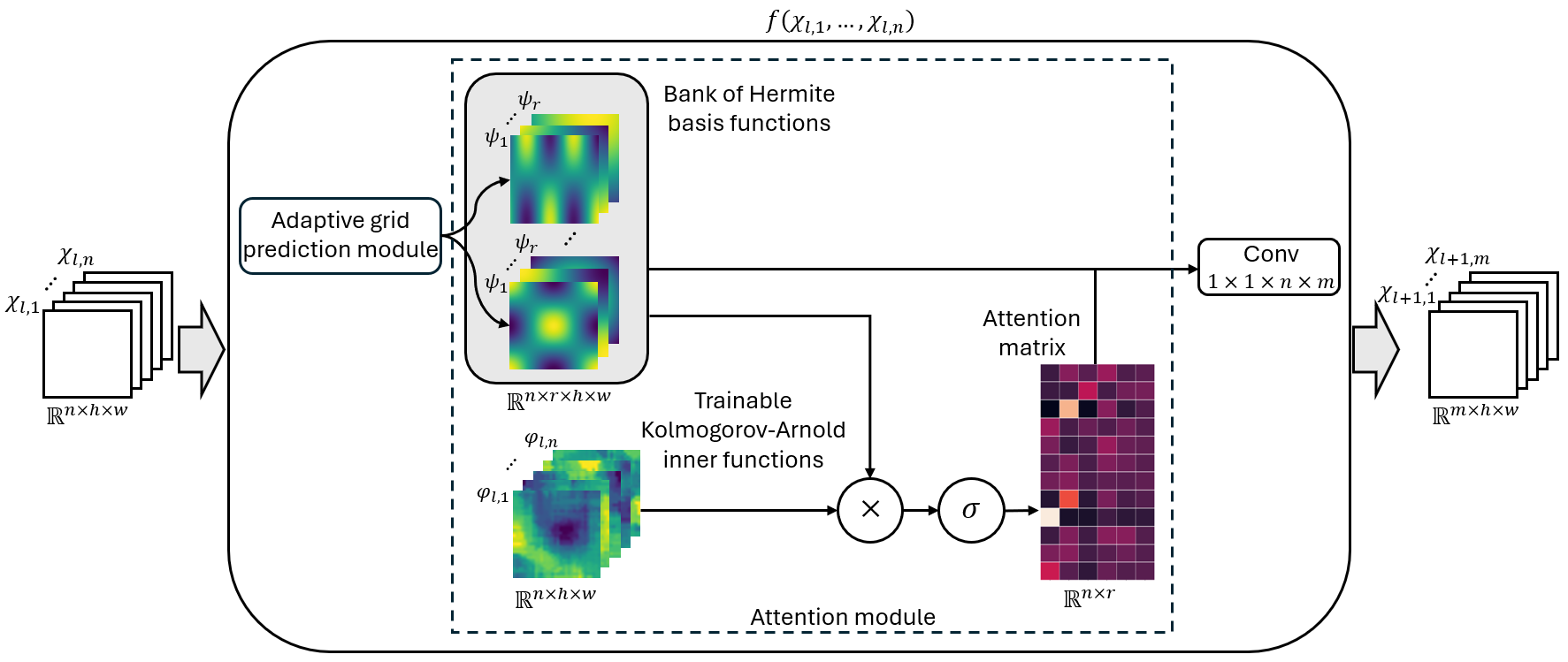}
\caption{Architecture of Functional Kolmogorov-Arnold Network (FunKAN).
The network implements trainable inner functions $\{ \varphi_{l,i} \}_{i=1}^{n}$
through Fourier decompositions over the basis Hermite functions  $\{ \psi_{k} \}_{k=1}^{r}$,
where each function is visualized on $h \times w$ spatial grid matching the input feature dimensions, 
and each decomposition is defined by normalized Fourier coefficients stored in the rows of the attention matrix.}
\label{fig_arch}
\end{figure*}
The architecture of our functional Kolmogorov-Arnold network is shown in Fig.~\ref{fig_arch}.
It derives from the following proposed hypothetical functional generalization of the Kolmogorov-Arnold theorem,
which we formally state:
\begin{theorem}
\label{mytheorem}
If $f$ is a continuous functional on $H^{n}$,
then it can be represented as a composition of
linear continuous functionals from the dual space $H^{*}$,
continuous functions of a single variable
and the binary operation of addition:
\[ f(\chi_{1},  ..., \chi_{n}) \leadsto \sum_{j} \zeta_{j} (\sum_{i} \varphi_{ji}(\chi_{i})), \]
where $H$ is a Hilbert space,
$\chi_{i} \in H$, 
$\varphi_{ji} \in H^{*}$ and
$\zeta_{j}$: $\mathbb{R} \rightarrow \mathbb{R}$.
\end{theorem}

While formally unproven, the proposed functional extension of the Kolmogorov-Arnold theorem hypothesizes
that continuous operators on $H^{n}$, modeling activation mappings between feature spaces,
may be approximated by functionals from the dual space $H^{*}$.

In contrast to the original KAN~\cite{liu2024kan}, such framework is \textit{naturally} adapted 
into deep image processing frameworks by construction,
considering each 2D feature map $\chi_{l, i}$ as a function from a Hilbert space, discretized over $h \times w$ spatial grid,
and computing the next-layer feature map $\chi_{l+1, j}$ as:
\begin{equation} \label{eq:funkan_1}
\chi_{l+1,j} = \sum_{i=1}^{n} \varphi_{l, ji} (\chi_{l, i}),
\end{equation}
where $\varphi_{l,ji} \in H^{*}$.

By invoking the Riesz representation theorem,
which establishes the isomorphism between a Hilbert space $H$ and its dual space $H^{*}$,
the equation \eqref{eq:funkan_1} simplifies to:
\begin{equation} \label{eq:funkan_2}
\chi_{l+1,j} = \sum_{i=1}^{n} \varphi_{l,ji},
\end{equation}
where $\varphi_{l,ji} \in H$, each responding to its $\chi_{l, i} \in H$.

To ensure a differentiable parameterization of the inner functions,
we adopt a constructive approach by representing each inner function $\varphi_{l,ji}$
via its Fourier expansion over the first $r$ Hermite functions.
This spectral decomposition allows us to retain the most informative modes, 
mirroring the frequency-truncation strategy employed in Fourier neural operator (FNO)~\cite{li2023fourier}:
\begin{equation} \label{eq:funkan_2_1}
\varphi_{l,ji} \leadsto \sum_{k=1}^{r} \inp{\varphi_{l,ji}}{\psi_{k}} \psi_{k}.
\end{equation}

By factoring the $j$-index outward, this representation naturally induces a convolution with $1 \times 1$ kernel, defined by weights $\{ \theta_{l,j} \}_{j=1}^{m}$,
yielding the overall computational form:
\begin{equation} \label{eq:funkan_3}
\chi_{l+1,j} = \sum_{i=1}^{n} \theta_{l,j} (\sum_{k=1}^{r} \inp{\varphi_{l,i}}{\psi_{k}} \psi_{k} ).
\end{equation}

The coefficients $c_{l,ik} = \inp{\varphi_{l,i}}{\psi_{k}}$ are stored in the interpretable attention matrix
$A_{l} = \{ c_{l,ik}\} \in \mathbb{R}^{n \times r}$.

\begin{figure}
\centering
\includegraphics[width=0.75\columnwidth]{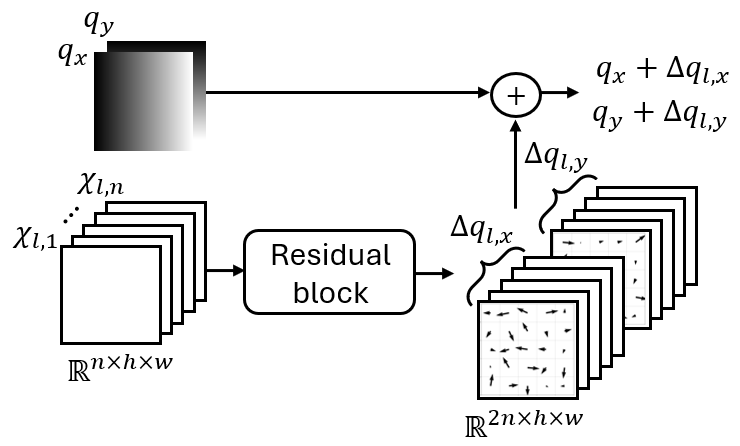}
\caption{Architecture of the adaptive grid prediction module,
illustrating learned spatial deformation through residual network-generated offset tensors $\Delta q_{l,x}$, $\Delta q_{l,y}$.
These predicted offsets are combined additively with a broadcasted uniform reference grid $\{q_x, q_y\}$
to produce the deformed sampling grid for the basis Hermite functions evaluation.}
\label{fig_offset}
\end{figure}
Inspired by state-of-the-art implicit architectures~\cite{agro2024uno},
the evaluation of the basis Hermite functions is performed on a dynamically adapted grid,
where the spatial coordinates are deformed
by a learned vector field $\Delta q_{l} = \{ \Delta q_{l,x}, \Delta q_{l,y} \}$,
generated through a residual block~\cite{he2016deep} (see Fig.~\ref{fig_offset}).
Formally, the grid deformation is computed as $q + \Delta q_{l}$,
where $q = \{ q_{x}, q_{y} \}$ is a uniform grid.

As shown in Fig.~\ref{fig_offset}, the residual block generates spatial offset tensors:
$\Delta q_{l,x} \in \mathbb{R}^{n \times h \times w}$ and 
$\Delta q_{l,y} \in \mathbb{R}^{n \times h \times w}$.
These offsets are then added to a uniform grid,
yielding the deformed sampling coordinates used for the basis Hermite functions evaluation.
Consistent with the pre-activation approach~\cite{duta2021improved},
the residual block processes activations through batch normalization~\cite{balestriero2022batch} and ReLU
before the subsequent convolution:
\begin{equation} \label{eq:offset_nonlinear1}
\Delta q_{l} = \mathcal{W}_{l,0} \circ BN(\chi_{l}) + \mathcal{F}_{l}(\chi_{l}),
\end{equation}
\begin{equation} \label{eq:offset_nonlinear2}
\mathcal{F}_{l} = \mathcal{W}_{l,2} \circ ReLU\{BN(\mathcal{W}_{l,1} \circ ReLU\{BN(\chi_{l})\})\},
\end{equation}
where
$\Delta q_{l} \in \mathbb{R}^{2n \times h \times w}$ -- learned spatial offsets,
$BN$ -- batch normalization and
$\{ \mathcal{W}_{l,i} \}_{i=0}^{2}$ -- convolutional layers, maintaining the spatial resolution,
encapsulating the trainable parameters:
$\mathbf{w}_{l,0} \in \mathbb{R}^{3 \times 3 \times n \times 2n}$,
$\mathbf{w}_{l,1} \in \mathbb{R}^{3 \times 3 \times n \times  n}$,
$\mathbf{w}_{l,2} \in \mathbb{R}^{3 \times 3 \times n \times 2n}$ -- kernels,
$\mathbf{b}_{l,1} \in \mathbb{R}^{n}$,
$\mathbf{b}_{l,2} \in \mathbb{R}^{2n}$ -- biases.

While both pre- and post-activation residual architectures possess equivalent theoretical representational capacity,
empirical evidence demonstrates superior gradient propagation in pre-activation architectures~\cite{duta2021improved}.
Our implementation (Fig.~\ref{fig_offset}) ensures stable optimization through 
batch normalization layers
and skip connections.

\begin{figure}
\centering
\includegraphics[width=0.88\columnwidth]{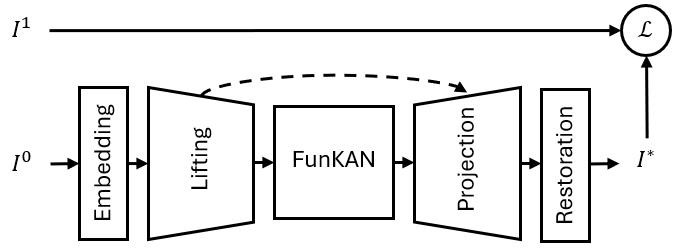}
\caption{Overview of FunKAN as a dual-purpose backbone
for medical image enhancement and segmentation.
The model processes an input image $I^{0}$,
being supervised by a target image $I^{1}$ via the loss function $\mathcal{L}(I^{*}, I^{1})$.}
\label{fig_overall}
\end{figure}
First, we leverage FunKAN as a backbone architecture (see Fig.~\ref{fig_overall})
for medical image enhancement.
In particular, the following setup is used for MRI enhancement:
\begin{enumerate}
    \item Embedding: $5 \times 5$ convolution, projecting an input image into 16-dimensional feature space.
    \item Lifting: $3 \times 3$ convolution, projecting ReLU pre-activated features into 32-dimensional feature space, maintaining the spatial resolution.
    \item Backbone: sequence of three FunKAN blocks, interconnected with skip connections in 32-dimensional feature space ($n=32$),
    encapsulating the spectral encoding of each inner function over the first six Hermite basis functions ($r=6$).
    \item Projection: $3 \times 3$ convolution, projecting ReLU pre-activated features into 16-dimensional feature space, maintaining the spatial resolution.
    \item Restoration: $1 \times 1$ convolution, projecting ReLU pre-activated features to the target color space.
\end{enumerate}
The model is trained in a supervised manner using mean squared error loss function:
\begin{equation} \label{eq:mse}
\mathcal{L}_{enh} = \frac{1}{N} \sum_{i=1}^{N} ||I^{*}_{i} - I^{1}_{i}||_{2}^{2},
\end{equation}
where $N$ -- batch size, equals to 8.

Second, we design U-FunKAN for medical segmentation,
incorporating FunKAN within the U-shaped segmentation framework (see Fig.~\ref{fig_overall}):
\begin{enumerate}
    \item Embedding: $3 \times 3$ convolution, projecting an input image into 16-dimensional feature space.
    \item Lifting: four consecutive U-Net-like encoder residual blocks with progressively increasing filter count:
    32 ($C_{1}$) $\rightarrow$  64 ($C_{2}$) $\rightarrow$ 128 ($C_{3}$) $\rightarrow$ 128, --
    each halving the spatial resolution through strided  $3 \times 3$ convolution.
    \item Backbone: sequence of three FunKAN blocks, interconnected with skip connections in 128-dimensional feature space ($n=128$),
    encapsulating the spectral encoding of each inner function over the first six Hermite basis functions ($r=6$).
    \item Projection: four consecutive U-Net-like decoder residual blocks with gradually decreasing filter count:
    128 ($C_{3}$) $\rightarrow$ 64 ($C_{2}$) $\rightarrow$ 32 ($C_{1}$) $\rightarrow$ 16, --
    each performing $\times 2$ upsampling via nearest-neighbor interpolation, followed by $3 \times 3$ convolution for feature refinement
    and skip connection from the corresponding lifting module.
    \item Restoration: $1 \times 1$ convolution, projecting ReLU pre-activated features to logits.
\end{enumerate}
The model is also trained in a supervised manner using the weighted combination of binary cross-entropy and dice loss:
\begin{equation} \label{eq:dice}
\mathcal{L}_{segm} = \frac{1}{N} \sum_{i=1}^{N} 0.1 \cdot CE(I^{*}_{i}, I^{1}_{i}) + Dice(I^{*}_{i}, I^{1}_{i}),
\end{equation}
where $N$ -- batch size, equals to 8.

The Hermite basis is utilized because of the inherent dual localization exhibited by Hermite functions,
a property stemming from their role as Fourier transform eigenfunctions~\cite{grunbaum1982eigenvectors}.
The number of basis functions ($r = 6$) is determined following the methodology of \cite{penkin2025adaptive},
which performed a grid search over three candidate bases—B-splines, Chebyshev polynomials and Hermite functions,
ultimately selecting six basis functions as optimal.

\section{Experiments}
We conducted a comprehensive evaluation of FunKAN
on four benchmark datasets,
selected to validate our method's robustness
across anatomical diversity,
encompassing neurological, oncological, histological and endoscopic structures,
and modality variations,
including MRI, ultrasound, histopathology and colonoscopy.

The experimental framework is implemented in Python 3.12 using PyTorch 2.5,
with all models trained and evaluated with full precision on NVIDIA RTX A6000 GPU.
The software stack employs PyTorch Lightning 2.5.1, CUDA 11.8 and cuDNN 9.
Computational reproducibility is ensured through the seeds setup
and YAML-based experiments management.
The models were trained from scratch till convergence using
Adam~\cite{diederik2014adam} stochastic optimization algorithm
($\beta_{1} = 0.9$,
$\beta_{2} = 0.999$,
$\varepsilon = 10^{-8}$)
with learning rate manual scheduling upon the scheme:
$10^{-4}$, $5 \cdot 10^{-5}$, $10^{-5}$.
To enhance generalization, MRI training dataset was augmented by Gaussian noise with $\sigma = 0.01$,
and segmentation training datasets were augmented
through random vertical/horizontal flips, rotations and transpositions, each applied with a probability of 0.5.

We measured MRI enhancement quality on IXI dataset, in terms of PSNR and TV,
and anatomical structures segmentation accuracy on BUSI, GlaS and CVC-ClinicDB datasets, in terms of IoU and F1 scores.
To ensure full reproducibility, we  released CSV files specifying the data splits for each dataset in our GitHub repository.

\begin{figure}
\centering
\includegraphics[width=\columnwidth]{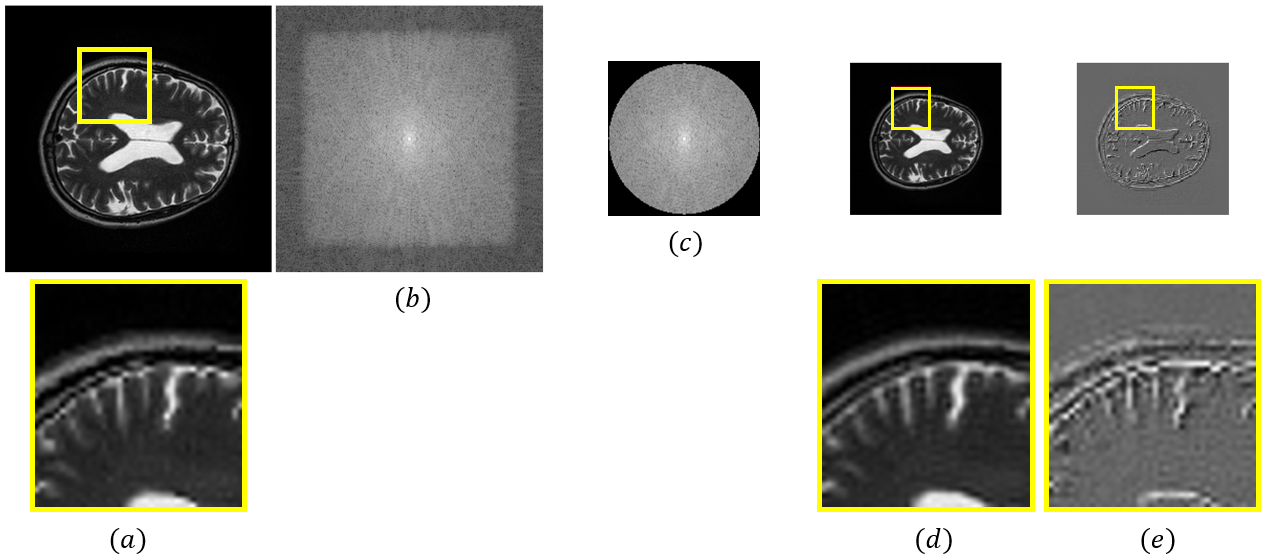}
\caption{
Synthetic pipeline for artifact-corrupted MRI samples generation from IXI dataset.
$(a)$ -- high-resolution image from IXI dataset,
$(b)$ -- Fourier spectrum of $(a)$,
$(c)$ -- Fourier spectrum truncation,
$(d)$ -- inverse Fourier transform of $(c)$,
         yielding low-resolution corrupted image,
$(e)$ -- residual image: $(d) - (a)$, -- demonstrating Gibbs ringing.
Yellow bounding boxes hightlight zoomed regions.}
\label{fig_mri_pipeline}
\end{figure}
\textbf{IXI dataset.}
IXI dataset~\cite{zhao2020gibbs} is a publicly available multi-modal neuroimaging dataset.
It comprises approximately 600 high-resolution MRI scans from healthy subjects collected using 1.5T and 3T scanners from Philips and GE Healthcare. 
It is composed of 581 T1, 578 T2 and 578 PD volumes, encoded in NIFTI format.
Firstly, the intersection of these volumes was taken, producing 577 volumes, which have all three modalities: T1, T2 and PD.
Then, the first 400 volumes were utilized to synthesize the training set,
the next 100 volumes to create the testing set
and the rest of the data to generate the validating set.
25 slices at both ends were discarded
and every tenth slice was obtained to produce a pair of artifact-corrupted and artifact-free MR images.
So, the training, validating and testing sets consist of 10427, 2016 and 2617 pairs, respectively,
produced via the protocol (see Fig.~\ref{fig_mri_pipeline}):
\begin{enumerate}
    \item Load high-resolution image of size $255 \times 255$.
    \item Apply Fourier transform.
    \item Crop central 25\% of the frequency domain. No zero-padding is applied~\cite{kellner2016gibbs}. 
    \item Apply inverse Fourier transform,
producing low-resolution Gibbs-corrupted image of size $145 \times 145$.
\end{enumerate}

\begin{table}
\setlength{\tabcolsep}{1mm}
\fontsize{9pt}{11pt}\selectfont
\centering
\begin{tabular}{lcccc}
\hline
\multirow{2}{*}{Methods} & \multicolumn{4}{c}{IXI}                                 \\ \cline{2-5} 
                         & PSNR $\uparrow$  & TV $\uparrow$    & Gflops $\downarrow$  & Params (M) $\downarrow$ \\ \hline
$I^{0}$                  & 31.33            & 1476.55          &  --                  & --                 \\
$I^{1}$                  & --               & 1255.40          &  --                  & --                 \\
Kellner                  & 31.09            & 1120.05          &  --                  & --                 \\
MLP                      & 37.96            & 1145.57          &    0.19              &  \textbf{0.01}     \\
KAN                      & 38.10            & 1161.63          &   \textbf{0.12}      &  0.04              \\ 
ChebyKAN                 & 38.01            & 1156.56          &   \textbf{0.12}      &  0.03              \\ 
HermiteKAN               & 38.04            & 1161.31          &   \textbf{0.12}      &  0.03              \\ \hline
FunKAN (Ours)            & \textbf{39.05}   & \textbf{1174.86} &    3.11              &  2.2               \\ \hline
\end{tabular}
\caption{Comparative analysis of MRI enhancement backbones within
the same convolutional architecture.
Results report average peak signal-to-noise ratio (PSNR) and total variation (TV) 
across 2617 test images ($145 \times 145$) from IXI dataset. $I^{0}$, $I^{1}$ denote artifact-corrupted and artifact-free images, respectively.}
\label{table0}
\end{table}

A quantitative comparison (see Table~\ref{table0}) of backbones,
within the same convolutional architecture, for MRI enhancement on IXI dataset
demonstrates that substituting MLP with KAN and subsequently retraining the entire model results in a modest PSNR improvement of 0.1 dB.
In contrast, FunKAN backbone
achieves a markedly higher gain, outperforming KANs by 1 dB,
owing to its inherent capacity to incorporate the geometric relations of visual data by design.
Performance is benchmarked against the Kellner deringing algorithm~\cite{kellner2016gibbs},
with superior restoration quality indicated by higher PSNR values and TV measures approaching the reference $I^{1}$.

\begin{figure}
\centering
\includegraphics[width=\columnwidth]{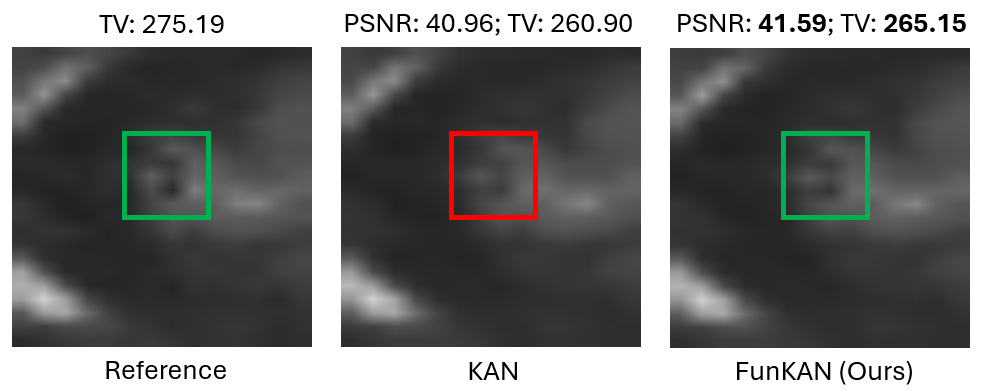}
\caption{
Visual comparison of MRI enhancement KAN and FunKAN backbones within the same convolutional architecture.
Bounding boxes highlight the challenging subtle anatomical structures and boundaries
with patch-wise peak signal-to-noise ratio (PSNR) and total variation (TV) values.}
\label{fig_ex}
\end{figure}
A qualitative comparison (see Fig.~\ref{fig_ex}) reveals KAN reconstruction to exhibit a moderate blur of fine structures,
whereas FunKAN preserves sharper edges and higher-frequency details.
Such fidelity is critical for clinical applications, as blurring can obscure diagnostically relevant features,
including early-stage pathologies or microstructural anomalies.
The improvements in PSNR and TV further corroborate FunKAN’s superior ability
to balance Gibbs ringing suppression with detail retention.

\begin{table*}
\fontsize{9pt}{11pt}\selectfont
\centering
\begin{tabular}{lcccccc}
\hline
\multirow{2}{*}{Methods}            & \multicolumn{2}{c}{BUSI}        & \multicolumn{2}{c}{GlaS}        & \multicolumn{2}{c}{CVC}         \\ \cline{2-7} 
                                    & IoU $\uparrow$ & F1 $\uparrow$  & IoU $\uparrow$ & F1 $\uparrow$  & IoU $\uparrow$ & F1 $\uparrow$  \\ \hline
U-Net~\cite{ronneberger2015u}       & 57.22±4.74     & 71.91±3.54     & 86.66±0.91     & 92.79±0.56     & 83.79±0.77     & 91.06±0.47     \\
Att-Unet~\cite{oktay2018attention}  & 55.18±3.61     & 70.22±2.88     & 86.84±1.19     & 92.89±0.65     & 84.52±0.51     & 91.46±0.25     \\
U-Net++~\cite{zhou2018unet++}       & 57.41±4.77     & 72.11±3.90     & 87.07±0.76     & 92.96±0.44     & 84.61±1.47     & 91.53±0.88     \\
U-NeXt~\cite{valanarasu2022unext}   & 59.06±1.03     & 73.08±1.32     & 84.51±0.37     & 91.55±0.23     & 74.83±0.24     & 85.36±0.17     \\
Rolling-UNet~\cite{liu2024rolling}  & 61.00±0.64     & 74.67±1.24     & 86.42±0.96     & 92.63±0.62     & 82.87±1.42     & 90.48±0.83     \\ 
U-Mamba~\cite{ma2024u}              & 61.81±3.24     & 75.55±3.01     & 87.01±0.39     & 93.02±0.24     & 84.79±0.58     & 91.63±0.39     \\ 
UKAGNet~\cite{drokin2024kolmogorov} & 63.45          & \textbf{77.64}          & 87.31          & 93.23          & 76.85          & 86.91 \\ 
U-KAN~\cite{li2025u}                & 63.38±2.83     & 76.40±2.90     & 87.64±0.32     & 93.37±0.16     & 85.05±0.53     & \textbf{91.88±0.29} \\ \hline
U-FunKAN (Ours)                    & \textbf{68.49±0.62}     & 77.37±0.58     & \textbf{88.02±0.24}     & \textbf{93.50±0.12}     & \textbf{85.93±0.72}     & 91.42±0.61     \\ \hline
\end{tabular}
\caption{Performance comparison of state-of-the-art segmentation models across three clinically distinct medical imaging scenarios.
Results report average intersection over union (IoU) and F1 scrores with standard deviation over three random runs.}
\label{table1}
\end{table*}

\begin{table*}
\fontsize{9pt}{11pt}\selectfont
\centering
\begin{tabular}{lcc}
\hline
Methods                            & Gflops $\downarrow$   & Params (M) $\downarrow$ \\ \hline
U-Net~\cite{ronneberger2015u}      & 524.2    & 34.53      \\
Att-Unet~\cite{oktay2018attention} & 533.1    & 34.9       \\
U-Net++~\cite{zhou2018unet++}      & 1109     & 36.6       \\
U-NeXt~\cite{valanarasu2022unext}  & 4.58     & \textbf{1.47} \\
Rolling-UNet~\cite{liu2024rolling} & 16.82    & 1.78       \\ 
U-Mamba~\cite{ma2024u}             & 2087     & 86.3       \\ 
U-KAN~\cite{li2025u}               & 14.02    & 6.35       \\ \hline
U-FunKAN (Ours)                    & \textbf{4.35}     & 3.6        \\ \hline
\end{tabular}
\caption{Efficiency comparison of floating-point operations (Gflops) and trainable parameters (Params) across state-of-the-art segmentation algorithms.
Results are measured using THOP Python tool for profiling PyTorch models.}
\label{table2}
\end{table*}

\textbf{BUSI dataset.}
BUSI dataset~\cite{al2020dataset} is a publicly available dataset for breast tumor segmentation
in ultrasound imaging.
It consists of 780 2D grayscale breast ultrasound images in PNG format,
collected from 600 female patients (aged 25–75) and categorized into three classes:
133 images with no visible tumors (normal),
437 images of non-cancerous lesions (benign)
and 210 images of confirmed cancerous tumors (malignant).
We utilized 647 benign and malignant images, resized to $256 \times 256$.

\textbf{GlaS dataset.}
GlaS dataset~\cite{valanarasu2021medical} is a widely used dataset,
specifically designed for gland segmentation.
It contains 165 Hematoxylin and Eosin (H\&E) stained histology RGB images.
Our study utilized 165 images, resized to $512 \times 512$.
Despite the predefined train-test division,
we re-partitioned them into training (80\%) and testing (20\%) subsets
using a randomized split with the seed 42.
Such approach ensures a fair comparison with competitors and the way we split BUSI and CVC datasets.

\textbf{CVC-ClinicDB dataset.}
CVC-ClinicDB dataset~\cite{bernal2015wm} is a dataset for polyp segmentation in colonoscopy images.
It contains 612 high-resolution colonoscopy RGB frames, extracted from 29 video sequences with
varied lighting conditions, specular reflections and mucosal textures.
All images were resized to $256 \times 256$.

U-FunKAN achieves state-of-the-art segmentation accuracy in terms of IoU
across all three medical imaging datasets (see Table~\ref{table1}), also being the most efficient algorithm in terms of Gflops (see Table~\ref{table2}).
Table~\ref{table1} reports the performance metrics, averaged over the last fifty epochs,
with uncertainty estimations, derived from three independent training runs with the seeds: 50, 100 and 150.
In terms of F1 score, U-FunKAN attains the highest quality on GlaS datasets,
while minor underperforming on BUSI and CVC datasets
compared to UKAGNet~\cite{drokin2024kolmogorov} and U-KAN~\cite{li2025u}, respectively,
which requires $\times 3$ greater computational complexity (in Gflops)
and $\times 1.7$ more parameters than our approach.

The comparative analysis of U-FunKAN model variants with different channel settings is reported in Table~\ref{table3}.
The proposed channel setting: 32 ($C_{1}$) $\rightarrow$  64 ($C_{2}$) $\rightarrow$ 128 ($C_{3}$), --
achieves an optimal balance between computational efficiency and segmentation performance, yielding state-of-the-art results.
While deeper configurations can further improve accuracy, they incur a significant computational overhead.

\begin{table}
\setlength{\tabcolsep}{1mm}
\fontsize{9pt}{11pt}\selectfont
\centering
\begin{tabular}{ccc|cccc}
\hline
\multicolumn{3}{c|}{U-FunKAN} & \multicolumn{4}{c}{BUSI} \\ \hline
$C_1$         & $C_2$            & $C_3$             & IoU $\uparrow$ & F1 $\uparrow$  & Gflops $\downarrow$ & Params (M) $\downarrow$  \\ \hline
\textbf{32}   & \textbf{64}      & \textbf{128}      & 69.11          & 77.95          & \textbf{4.35}       & \textbf{3.6}   \\ \hline
64            & 96               & 128               & 69.94          & 78.42          & 10.84               & 4.1            \\
128           & 160              & 256               & 69.49          & 78.39          & 40.42               & 15.7           \\ 
256           & 320              & 512               & \textbf{70.62} & \textbf{79.31} & 161.43              & 62.4           \\ \hline
\end{tabular}
\caption{Ablation study on impact of the channel scaling in U-FunKAN on segmentation performance and efficiency.}
\label{table3}
\end{table}

\section{Conclusion}
In this paper we proposed FunKAN -- a novel state-of-the-art neural framework for medical image enhancement and segmentation.
The proposed generalization of the Kolmogorov-Arnold representation theorem to functional spaces
remains empirically validated rather than formally proven, however, our extensive experiments demonstrate its superiority
over competitors.
Specifically, we explored FunKAN as a backbone for MRI enhancement
and introduced state-of-the-art U-FunKAN, a segmentation variant, applied to
breast tumor detection in ultrasound,
gland segmentation in histology and polyp identification in colonoscopy videos.
Our code will be publicly available in case of the paper acceptance.

\bibliography{aaai2026}

\begin{thebibliography}{38}
\providecommand{\natexlab}[1]{#1}

\bibitem[{Abhisheka, Biswas, and Purkayastha(2023)}]{abhisheka2023comprehensive}
Abhisheka, B.; Biswas, S.~K.; and Purkayastha, B. 2023.
\newblock A Comprehensive Review on Breast Cancer Detection, Classification and Segmentation using Deep Learning.
\newblock \emph{Archives of Computational Methods in Engineering}, 30(8): 5023--5052.

\bibitem[{Agro et~al.(2024)Agro, Sykora, Casas, Gilles, and Urtasun}]{agro2024uno}
Agro, B.; Sykora, Q.; Casas, S.; Gilles, T.; and Urtasun, R. 2024.
\newblock UnO: Unsupervised Occupancy Fields for Perception and Forecasting.
\newblock In \emph{Proceedings of the IEEE/CVF Conference on Computer Vision and Pattern Recognition}, 14487--14496.

\bibitem[{Al-Dhabyani et~al.(2020)Al-Dhabyani, Gomaa, Khaled, and Fahmy}]{al2020dataset}
Al-Dhabyani, W.; Gomaa, M.; Khaled, H.; and Fahmy, A. 2020.
\newblock Dataset of Breast Ultrasound Images.
\newblock \emph{Data in brief}, 28: 104863.

\bibitem[{Balestriero and Baraniuk(2022)}]{balestriero2022batch}
Balestriero, R.; and Baraniuk, R.~G. 2022.
\newblock Batch Normalization Explained.
\newblock \emph{arXiv preprint arXiv:2209.14778}.

\bibitem[{Bernal et~al.(2015)Bernal, S{\'a}nchez, Fern{\'a}ndez-Esparrach, Gil, Rodr{\'\i}guez, and Vilari{\~n}o}]{bernal2015wm}
Bernal, J.; S{\'a}nchez, F.~J.; Fern{\'a}ndez-Esparrach, G.; Gil, D.; Rodr{\'\i}guez, C.; and Vilari{\~n}o, F. 2015.
\newblock WM-DOVA Maps for Accurate Polyp Highlighting in Colonoscopy: Validation vs. Saliency Maps from Physicians.
\newblock \emph{Computerized medical imaging and graphics}, 43: 99--111.

\bibitem[{Borys et~al.(2023)Borys, Schmitt, Nauta, Seifert, Kr{\"a}mer, Friedrich, and Nensa}]{borys2023explainable}
Borys, K.; Schmitt, Y.~A.; Nauta, M.; Seifert, C.; Kr{\"a}mer, N.; Friedrich, C.~M.; and Nensa, F. 2023.
\newblock Explainable AI in Medical Imaging: An Overview for Clinical Practitioners -- Beyond Saliency-based XAI Approaches.
\newblock \emph{European journal of radiology}, 162: 110786.

\bibitem[{Cybenko(1989)}]{cybenko1989approximation}
Cybenko, G. 1989.
\newblock Approximation by Superpositions of a Sigmoidal Function.
\newblock \emph{Mathematics of control, signals and systems}, 2(4): 303--314.

\bibitem[{Diederik(2014)}]{diederik2014adam}
Diederik, P.~K. 2014.
\newblock Adam: A Method for Stochastic Optimization.
\newblock \emph{(No Title)}.

\bibitem[{Drokin(2024)}]{drokin2024kolmogorov}
Drokin, I. 2024.
\newblock Kolmogorov-Arnold Convolutions: Design Principles and Empirical Studies.
\newblock \emph{arXiv preprint arXiv:2407.01092}.

\bibitem[{Duta et~al.(2021)Duta, Liu, Zhu, and Shao}]{duta2021improved}
Duta, I.~C.; Liu, L.; Zhu, F.; and Shao, L. 2021.
\newblock Improved Residual Networks for Image and Video Recognition.
\newblock In \emph{2020 25th International Conference on Pattern Recognition (ICPR)}, 9415--9422. IEEE.

\bibitem[{Epstein(2007)}]{epstein2007introduction}
Epstein, C.~L. 2007.
\newblock \emph{Introduction to the Mathematics of Medical Imaging}.
\newblock SIAM.

\bibitem[{Gr{\"u}nbaum(1982)}]{grunbaum1982eigenvectors}
Gr{\"u}nbaum, F.~A. 1982.
\newblock The Eigenvectors of the Discrete Fourier Transform: A Version of the Hermite Functions.
\newblock \emph{Journal of Mathematical Analysis and Applications}, 88(2): 355--363.

\bibitem[{He et~al.(2016)He, Zhang, Ren, and Sun}]{he2016deep}
He, K.; Zhang, X.; Ren, S.; and Sun, J. 2016.
\newblock Deep Residual Learning for Image Recognition.
\newblock In \emph{Proceedings of the IEEE conference on computer vision and pattern recognition}, 770--778.

\bibitem[{Hewitt and Hewitt(1979)}]{wilton1928gibbs}
Hewitt, E.; and Hewitt, R.~E. 1979.
\newblock The Gibbs-Wilbraham Phenomenon: An Episode in Fourier Analysis.
\newblock \emph{Archive for history of Exact Sciences}, 129--160.

\bibitem[{Kadhim, Khan, and Mishra(2022)}]{kadhim2022deep}
Kadhim, Y.~A.; Khan, M.~U.; and Mishra, A. 2022.
\newblock Deep learning-based Computer-Aided Diagnosis (CAD): Applications for Medical Image Datasets.
\newblock \emph{Sensors}, 22(22): 8999.

\bibitem[{Kellner et~al.(2016)Kellner, Dhital, Kiselev, and Reisert}]{kellner2016gibbs}
Kellner, E.; Dhital, B.; Kiselev, V.~G.; and Reisert, M. 2016.
\newblock Gibbs-ringing Artifact Removal based on Local Subvoxel-shifts.
\newblock \emph{Magnetic resonance in medicine}, 76(5): 1574--1581.

\bibitem[{Kolmogorov(1957)}]{kolmogorov1957representations}
Kolmogorov, A.~N. 1957.
\newblock On the Representations of Continuous Functions of Many Variables by Superposition of Continuous Functions of One Variable and Addition.
\newblock In \emph{Dokl. Akad. Nauk USSR}, volume 114, 953--956.

\bibitem[{Li et~al.(2025)Li, Liu, Li, Wang, Liu, Liu, Chen, and Yuan}]{li2025u}
Li, C.; Liu, X.; Li, W.; Wang, C.; Liu, H.; Liu, Y.; Chen, Z.; and Yuan, Y. 2025.
\newblock U-KAN Makes Strong Backbone for Medical Image Segmentation and Generation.
\newblock \emph{Proceedings of the AAAI Conference on Artificial Intelligence}, 39(5): 4652--4660.

\bibitem[{Li(2024)}]{li2024kolmogorov}
Li, Z. 2024.
\newblock Kolmogorov-Arnold Networks are Radial Basis Function Networks.
\newblock \emph{arXiv preprint arXiv:2405.06721}.

\bibitem[{Li et~al.(2023)Li, Huang, Liu, and Anandkumar}]{li2023fourier}
Li, Z.; Huang, D.~Z.; Liu, B.; and Anandkumar, A. 2023.
\newblock Fourier Neural Operator with Learned Deformations for PDEs on General Geometries.
\newblock \emph{Journal of Machine Learning Research}, 24(388): 1--26.

\bibitem[{Liu et~al.(2024{\natexlab{a}})Liu, Zhu, Liu, Yu, Chen, and Gao}]{liu2024rolling}
Liu, Y.; Zhu, H.; Liu, M.; Yu, H.; Chen, Z.; and Gao, J. 2024{\natexlab{a}}.
\newblock Rolling-UNet: Revitalizing MLP’s Ability to Efficiently Extract Long-distance Dependencies for Medical Image Segmentation.
\newblock In \emph{Proceedings of the AAAI conference on artificial intelligence}, volume~38, 3819--3827.

\bibitem[{Liu et~al.(2024{\natexlab{b}})Liu, Wang, Vaidya, Ruehle, Halverson, Solja{\v{c}}i{\'c}, Hou, and Tegmark}]{liu2024kan}
Liu, Z.; Wang, Y.; Vaidya, S.; Ruehle, F.; Halverson, J.; Solja{\v{c}}i{\'c}, M.; Hou, T.~Y.; and Tegmark, M. 2024{\natexlab{b}}.
\newblock KAN: Kolmogorov-Arnold Networks.
\newblock \emph{arXiv preprint arXiv:2404.19756}.

\bibitem[{Ma, Li, and Wang(2024)}]{ma2024u}
Ma, J.; Li, F.; and Wang, B. 2024.
\newblock U-Mamba: Enhancing Long-range Dependency for Biomedical Image Segmentation.
\newblock \emph{arXiv preprint arXiv:2401.04722}.

\bibitem[{Mallat(1999)}]{mallat1999wavelet}
Mallat, S. 1999.
\newblock \emph{A Wavelet Tour of Signal Processing}.
\newblock Elsevier.

\bibitem[{Oktay et~al.(2018)Oktay, Schlemper, Folgoc, Lee, Heinrich, Misawa, Mori, McDonagh, Hammerla, Kainz et~al.}]{oktay2018attention}
Oktay, O.; Schlemper, J.; Folgoc, L.~L.; Lee, M.; Heinrich, M.; Misawa, K.; Mori, K.; McDonagh, S.; Hammerla, N.~Y.; Kainz, B.; et~al. 2018.
\newblock Attention U-Net: Learning Where to Look for the Pancreas.
\newblock \emph{arXiv preprint arXiv:1804.03999}.

\bibitem[{Penkin and Krylov(2023)}]{penkin2023medical}
Penkin, M.; and Krylov, A. 2023.
\newblock Medical Image Joint Deringing and Denoising using Fourier Neural Operator.
\newblock In \emph{Proceedings of the 2023 8th International Conference on Biomedical Imaging, Signal Processing}, 40--45.

\bibitem[{Penkin and Krylov(2025)}]{penkin2025adaptive}
Penkin, M.; and Krylov, A. 2025.
\newblock Adaptive Method for Selecting Basis Functions in Kolmogorov-Arnold Networks for Magnetic Resonance Image Enhancement.
\newblock \emph{Programming and Computer Software}, 51(3): 167--172.

\bibitem[{Penkin, Krylov, and Khvostikov(2021)}]{penkin2021hybrid}
Penkin, M.~A.; Krylov, A.~S.; and Khvostikov, A.~V. 2021.
\newblock Hybrid Method for Gibbs-ringing Artifact Suppression in Magnetic Resonance Images.
\newblock \emph{Programming and Computer Software}, 47(3): 207--214.

\bibitem[{Rahaman et~al.(2019)Rahaman, Baratin, Arpit, Draxler, Lin, Hamprecht, Bengio, and Courville}]{rahaman2019spectral}
Rahaman, N.; Baratin, A.; Arpit, D.; Draxler, F.; Lin, M.; Hamprecht, F.; Bengio, Y.; and Courville, A. 2019.
\newblock On the Spectral Bias of Neural Networks.
\newblock In \emph{International conference on machine learning}, 5301--5310. PMLR.

\bibitem[{Ronneberger, Fischer, and Brox(2015)}]{ronneberger2015u}
Ronneberger, O.; Fischer, P.; and Brox, T. 2015.
\newblock U-Net: Convolutional Networks for Biomedical Image Segmentation.
\newblock In \emph{International Conference on Medical image computing and computer-assisted intervention}, 234--241. Springer.

\bibitem[{Seydi(2024)}]{seydi2024exploring}
Seydi, S.~T. 2024.
\newblock Exploring the Potential of Polynomial Basis Functions in Kolmogorov-Arnold Networks: A Comparative Study of Different Groups of Polynomials.
\newblock \emph{arXiv preprint arXiv:2406.02583}.

\bibitem[{SS et~al.(2024)SS, AR, KP et~al.}]{ss2024chebyshev}
SS, S.; AR, K.; KP, A.; et~al. 2024.
\newblock Chebyshev Polynomial-based Kolmogorov-Arnold Networks: An Efficient Architecture for Nonlinear Function Approximation.
\newblock \emph{arXiv preprint arXiv:2405.07200}.

\bibitem[{Valanarasu et~al.(2021)Valanarasu, Oza, Hacihaliloglu, and Patel}]{valanarasu2021medical}
Valanarasu, J. M.~J.; Oza, P.; Hacihaliloglu, I.; and Patel, V.~M. 2021.
\newblock Medical Transformer: Gated Axial-attention for Medical Image Segmentation.
\newblock In \emph{International conference on medical image computing and computer-assisted intervention}, 36--46. Springer.

\bibitem[{Valanarasu and Patel(2022)}]{valanarasu2022unext}
Valanarasu, J. M.~J.; and Patel, V.~M. 2022.
\newblock UNeXt: MLP-based Rapid Medical Image Segmentation Network.
\newblock In \emph{International conference on medical image computing and computer-assisted intervention}, 23--33. Springer.

\bibitem[{Veraart et~al.(2016)Veraart, Fieremans, Jelescu, Knoll, and Novikov}]{veraart2016gibbs}
Veraart, J.; Fieremans, E.; Jelescu, I.~O.; Knoll, F.; and Novikov, D.~S. 2016.
\newblock Gibbs Ringing in Diffusion MRI.
\newblock \emph{Magnetic resonance in medicine}, 76(1): 301--314.

\bibitem[{Yang et~al.(2025)Yang, Zhang, Luo, Lu, and Shen}]{yang2025medkan}
Yang, Z.; Zhang, J.; Luo, X.; Lu, Z.; and Shen, L. 2025.
\newblock MedKAN: An Advanced Kolmogorov-Arnold Network for Medical Image Classification.
\newblock \emph{arXiv preprint arXiv:2502.18416}.

\bibitem[{Zhao et~al.(2020)Zhao, Zhang, Zhou, Bian, Zhang, and Zou}]{zhao2020gibbs}
Zhao, X.; Zhang, H.; Zhou, Y.; Bian, W.; Zhang, T.; and Zou, X. 2020.
\newblock Gibbs-ringing Artifact Suppression with Knowledge Transfer from Natural Images to MR Images.
\newblock \emph{Multimedia Tools and Applications}, 79(45): 33711--33733.

\bibitem[{Zhou et~al.(2018)Zhou, Rahman~Siddiquee, Tajbakhsh, and Liang}]{zhou2018unet++}
Zhou, Z.; Rahman~Siddiquee, M.~M.; Tajbakhsh, N.; and Liang, J. 2018.
\newblock UNet++: A Nested U-Net Architecture for Medical Image Segmentation.
\newblock In \emph{International workshop on deep learning in medical image analysis}, 3--11. Springer.

\end{thebibliography}
\end{document}